RESEARCH

# MDU-Net: multi-scale densely connected U-Net for biomedical image segmentation

Jiawei Zhang[1,2,3,4*], Yanchun Zhang[1,3,4], Yuzhen Jin[5], Jilan Xu[5] and Xiaowei Xu[2]

## Abstract

Biomedical image segmentation plays a central role in quantitative analysis, clinical diagnosis, and medical intervention. In the light of the fully convolutional networks (FCN) and U-Net, deep convolutional networks (DNNs) have made significant contributions to biomedical image segmentation applications. In this paper, we propose three different multi-scale dense connections (MDC) for the encoder, the decoder of U-shaped architectures, and across them. Based on three dense connections, we propose a multi-scale densely connected U-Net (MDU-Net) for biomedical image segmentation. MDU-Net directly fuses the neighboring feature maps with different scales from both higher layers and lower layers to strengthen feature propagation in the current layer. Multi-scale dense connections, which contain shorter connections between layers close to the input and output, also make a much deeper U-Net possible. Besides, we introduce quantization to alleviate the potential overfitting in dense connections, and further improve the segmentation performance. We evaluate our proposed model on the MICCAI 2015 Gland Segmentation (GlaS) dataset. The three MDC improve U-Net performance by up to 1.8% on test A and 3.5% on test B in the MICCAI Gland dataset. Meanwhile, the MDU-Net with quantization obviously improves the segmentation performance of original U-Net.

## Introduction

Biological structures play a central role in medical diagnosis, surgical planning, and treatments. Segmentation of the target tissue can indicate the patient's physical health, assist in diagnosing the severity of the patient's disease, and even deeply participate in the patient's surgical planning. However, it is difficult to obtain high-level pixel predictions in biomedical image segmentation, due to the diverse histological variation of targets in biomedical images. Meanwhile, medical image segmentation is often performed by experienced doctors traditionally, which is a time-consuming and expensive process. Therefore, accurate automatic medical image segmentation attracts people's attention and has wide application prospects.

Based on fully convolutional networks (FCN) and U-Net [1], deep convolutional networks (DNNs) have made significant improvements in biomedical image segmentation. Among them, using skip connections to connect different layers of the network to promote feature reuse and feature fusion is one of the hot topics. For example, U-Net employed skip connections, especially dense connections, to combine feature maps from the current layer with higher layer feature maps, which achieves competitive performance in maintaining fine-grained information. Recent researches on dense connections can be divided into two categories. (1) intra-block dense connections. It embeds the dense block to the traditional convolutional block such as FDU-Net [2]. In addition, cascaded stacked U-Nets also gain enough attention. CU-Net [3] performed dense connections of the same level among multiple U-Nets. However, these works fail to consider the utilization of feature maps with different scales. As a consequence, they are substantially different from our work. (2) Inter-block dense connections. It means the current layer can fuse feature maps from the previous layer with different scales. For instance, MIMO-Net [4] took input images of different scales in the encoder unit. However,

*Correspondence: 17110240008@fudan.edu.cn
[1] The Department of New Networks, Peng Cheng Laboratory, Shenzhen, Guangdong, China
Full list of author information is available at the end of the article the feature maps are not actually reused. U-Net++ [5] fused higher resolution feature maps in the decoder unit but it involves massively computational costs due to a large number of intermediate convolutions. In U-Net++, the current layer can only fuse the feature maps from higher layers.

In this paper, we introduce the multi-scale dense connections (MDC), which directly resize the learned features with multiple sizes to the same resolution as the features in the current layer and fuse them for better feature representation. We use 1×1 convolutional layers to adjust the number of channels the same as before. As illustrated in Fig. 1, the whole operation involves small extra parameters. As far as we are concerned, we are the first to explore directly fusing deep semantic and coarse-grained feature maps from higher layers and low-level, fine-grained feature maps from lower layers to boost the segmentation performance of the neural network. We also systematically analyze the impact of different kinds of densely connected structures. The experimental results show that fusing feature maps from higher and lower layers simultaneously can achieves higher precision. The contributions of our works are summarized as follows, (1) We propose three different dense connections for the encoder, and the decoder of the U-shape architectures and across them to fuse multi-scale features. (2) We explore the effectiveness of different dense connections in detail and proposed a novel multi-scale densely connected U-Net (MDU-Net) architecture for biomedical image segmentation. (3) We conduct detailed experiments and analyses of MDU-Net. The experimental results demonstrate that MDU-Net obviously improves the segmentation performance of original U-Net.

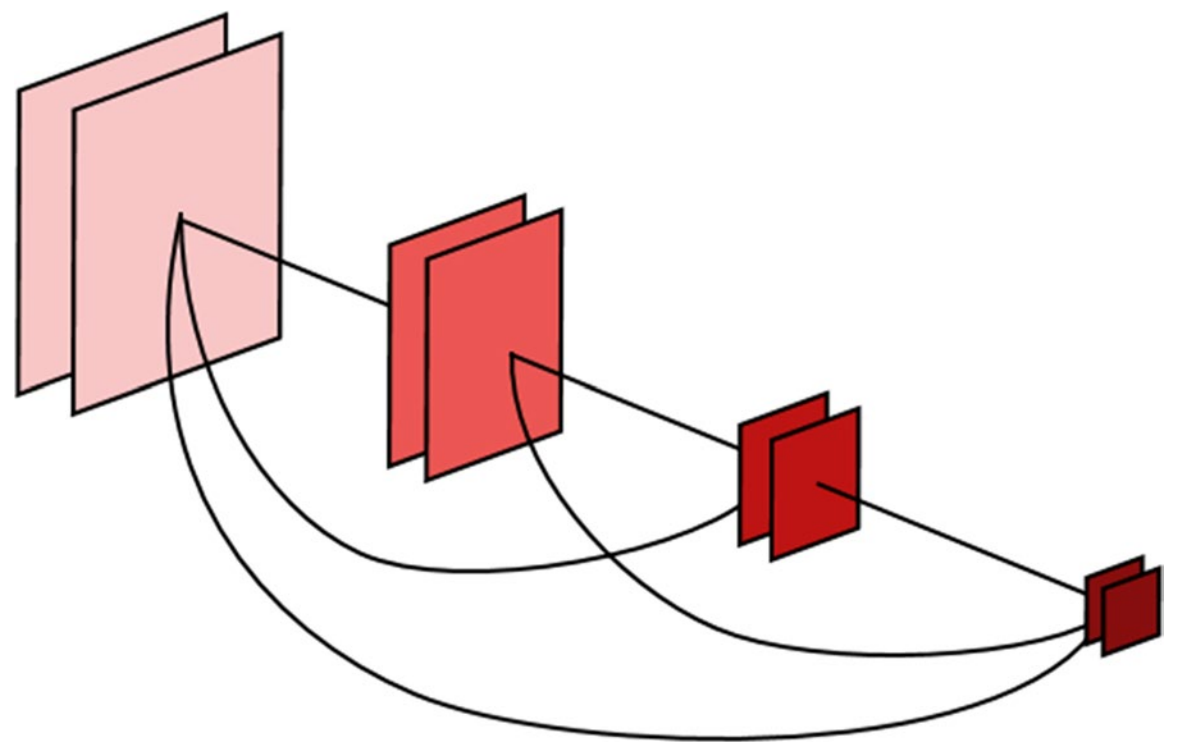

**Fig. 1** An example of the multi-scale dense encoder

## Related work

In this section, we introduce recent approaches toward U-Net architecture, dense connections, multi-scale feature aggregation, and biomedical image segmentation methods.

### U-Net architecture

Recently, in the light of the U-Net, models are always designed as encoder-decoder architectures to retrieve high resolution from low-resolution representations of the image to obtain the optimal segmentation performance. Other than image segmentation, a variety of tasks are involved in U-Net based architecture [6]. Stacked U-Nets [7] iteratively fused multi-scale features without changing the resolutions. To deal with human pose estimation tasks [8–10], stacked modified U-Nets which captured both the top-down and bottom-up features as a whole. Reference [11] additionally employed multi-path refinement and global convolutional blocks respectively between the encoder and decoder. The classification and localization problems are solved simultaneously during the successive down-sampling and up-sampling operations in U-Net. Furthermore, Reference [12] roposed a new lung parenchyma segmentation network, which introduces a dual U-Net with the utilization of the characteristics of different lung regions.

### Dense connections

Recently, the exploration of both the depth and the width of the network architecture has been a focused study. Approaches toward the wider network begin with [13, 14], which introduced 'Inception Module' by concatenating feature maps to approximate sparse structure. Moreover, residual network [15, 16] alleviated the vanishing gradient problem by summing up a shortcut connection with the residual function. Recent methods such as PSP-Net [17] and RefineNet [11] applied residual architecture more frequently as feature extractors in dense prediction tasks [18]. Combined U-Net with residual network and proved to skip connection effective in biomedical image segmentation. Additionally, to improve the representational power without increasing the depth and width of the network, Reference [19] proposed a typical structure of dense connections. In a dense block, each output of the convolution unit contributes to all the subsequent units as input through concatenation. With substantially fewer parameters, the network enables feature reuse and better gradient flow and therefore yields extremely competitive results. In FC-DenseNet [20], they extended the DenseNet [19] by replacing each convolutional block in the downsampling path of FCN with the dense block which they referred to as transition up the module to deal

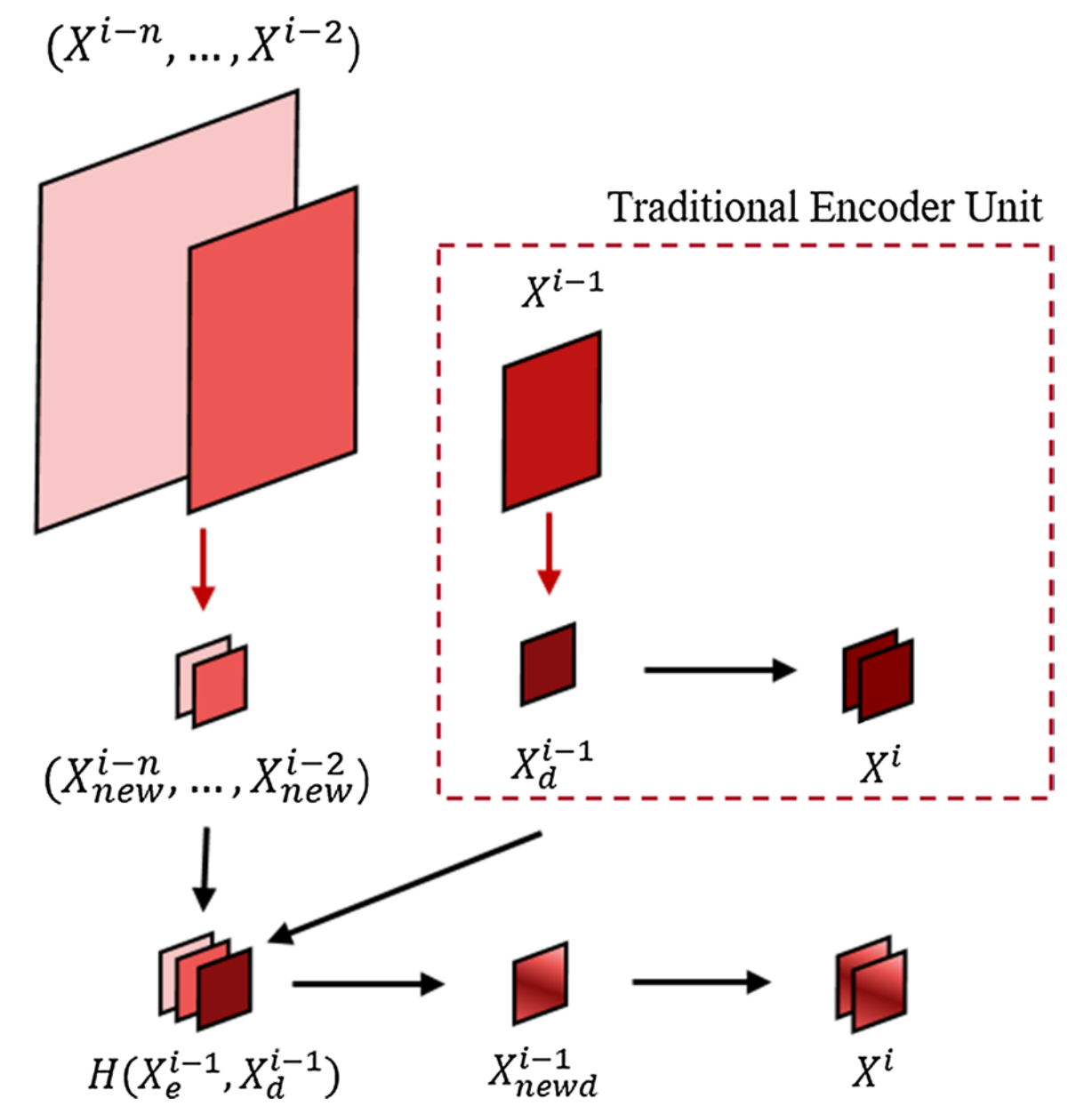

**Fig. 2** The comparison of the traditional encoder in U-Net and our dense encoder

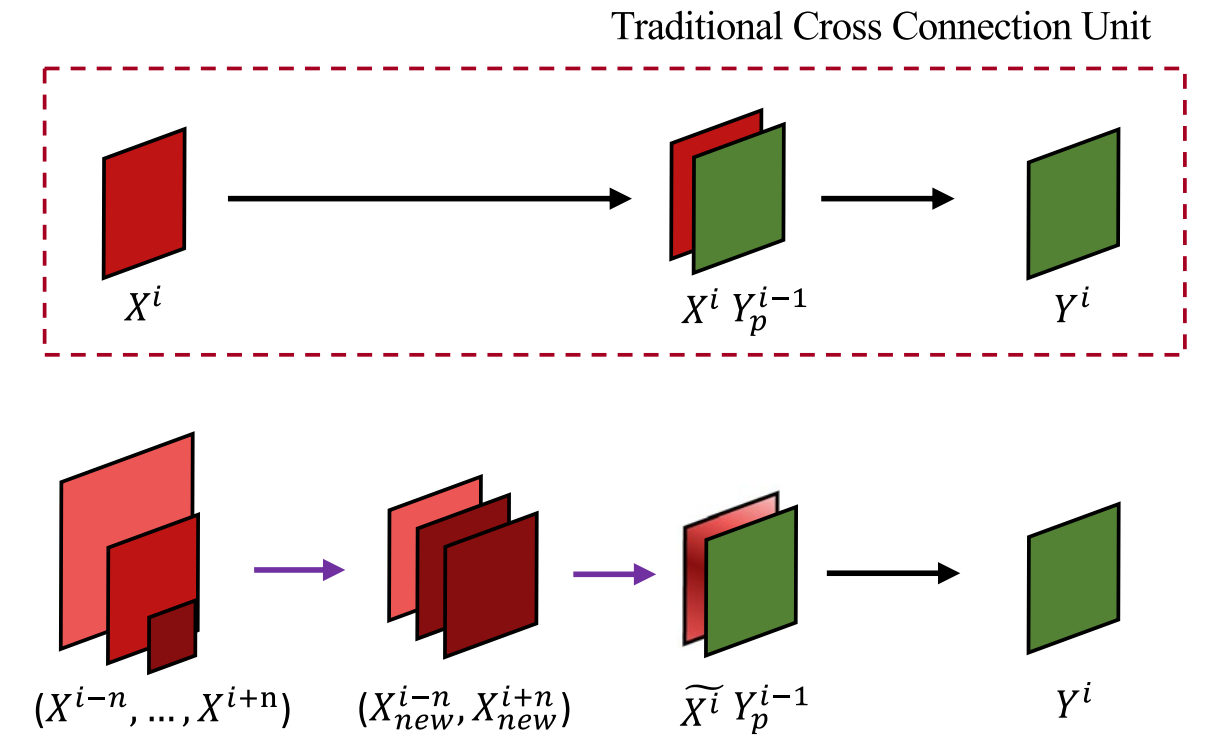

**Fig. 3** The comparison of the traditional connections across encoder and decoder in U-Net and our multi-scale dense connections across encoder and decoder

with semantic segmentation problems. Reference [21] further improved dense decoder blocks with feature-level long-range skip connections. With the cascaded architecture of single-pass, the network obtained surprising results with fewer computational costs on multi-scale works. The compact structure of dense connections integrates shortcut connection, feature reuse and implicit deep supervision while exhibiting no extra difficulties of optimization. Apart from directly adding dense connections in convolutional blocks, Reference [22] composed a denser scale sampling and denser pixel sampling in an atrous spatial pyramid pooling module [23]. Dense connections proved extraordinarily effective in biomedical image processing due to the limited amount of data. Reference [2] incorporated dense connectivity [24] within the encoder and decoder path. To address the spatial information of 3D input data, Reference [25] used 2D-Dense U-Net as an intra-slice feature extractor along with the hybrid feature fusion module to formulate end-to-end learning. Inspired by the previous literature, we generalize the dense connections to extend feature fusion and contextual information of various scales between the encoder and decoder.

### Multi-scale feature aggregation

Approaches towards the application of encoding multi-scale context information are widely explored. Other than the encoder-decoder structure discussed before, the construction of image pyramid [26, 27] is frequently used so that various scales of objects are obtained in the network. Dilated or atrous convolution [23, 28] deployed in parallel or cascaded expands the receptive fields while exhibiting no extra parameters. Further, ASPP [23] modified the atrous convolution in parallel within spatial pyramid pooling to efficiently capture features of an arbitrary scale. In particular, Dense-ASPP [22] stacked ASPP module in a denser manner. Beyond atrous convolution, deformable convolution [29] generalized the atrous convolution by boosting the spatial sampling locations. Besides, some research [30] constructed two auxiliary brunches in each block to produce and fuse coarse-to-fine context information to improve the overall performance. Recently, CMD-Net [31] employed NAS methods to explore the efficiency of the multi-scale connections, and construct a constrained multi-scale densely connected network, which achieves high performance with reduced computational cost. CMM-Net [32] developed an end-to-end neural network, which fuses the global contextual features with multiple scales in each level of the U-Net.

### Biomedical image segmentation

Previously, hand-crafted features containing morphological information are designed and traditional graph-based models are frequently used [33–36]. However, malignant subjects vary seriously in appearance and they are beyond the capacity of traditional methods. Therefore, deep learning methods have dominated biomedical image processing in recent years [37–40], especially in histological image analysis [39]. To relieve the effort of manual annotation, suggestive annotation [41] combined a fully convolutional network with active learning to select hard examples for further annotation. In addition, MIMO-Net [4] dealed with the variation of intense cell boundaries and sizes by exploiting multi-inputs and multi-outputs in the network. To this end, we propose a simple yet effective multi-scale connectivity pattern for biomedical image segmentation. Recently, GCSBA-Net [42] introduced three modules to model semantic correlation and maintain the multi-level aggregation on the spatial pyramid. Besides, no new U-Net (nnU-Net) [43] adjusts all hyperparameters according to the attributes of a given data set, without manual intervention, which is adaptive to many new datasets.

## Method

In this section, we first introduce three multi-scale densely connected blocks in the encoder, the decoder, and across them. Then we introduce the MDU-Net, which combines three multi-scale dense connections.

### Dense connections

In this section, we introduce three different dense connections for the encoder, and decoder of the U-shape architectures and across them. We use the U-Net as the base model, which is widely used in biomedical image segmentation. Firstly, we introduce the multi-scale dense connections in the encoder. We briefly look back at the basic structure of U-Net. A traditional encoder can be defined as the dotted rectangle in Fig. 2. $X^{i-1}$ and $X^i$ are the input and output of the current layer, respectively. $X_d^{i-1}$ is the output of $X^{i-1}$ after down-sample. Equations 1 and 2 describe the process.

$$X_d^{i-1} = D(X^{i-1}) \tag{1}$$

$$X^i = F(X_d^{i-1}) \tag{2}$$

As shown in Eq. 3, our multi-scale dense connections in encoder use $X_e^{i-1}$ to encode the feature maps $X_{new}^{(i-n)}, \cdots, X_{new}^{i-1}$, which are adjusted to the same size as $X_d^{i-1}$ from previous layer $I-n$ to layer $I-2$. Our method uses $X_{newd}^{i-1}$ instead of $X_d^{i-1}$, which is defined as Eq. 4. $X_{newd}^{(i-1)}$ fuses two feature maps $X_e^{i-1}$ and $X_d^{i-1}$. $H()$ represents the concatenation operation and conv1 $\times$ 1. The description of $n$ above refers to the number of current layers fuses ordered previous layer feature maps. the influence of the densely connected number $n$ will be discussed in "Quantitative results" section.

$$X_e^{i-1} = H(X_{new}^{(i-n)}, \cdots\cdots, X_{new}^{i-1}) \tag{3}$$

$$X_{newd}^{i-1} = H(X_e^{i-1}, X_d^{i-1}) \tag{4}$$

Specifically, each convolutional block is composed of two repeated cascaded structures of a conv 3 $\times$ 3, all of them follows by batch normalization and a ReLU activation function. Figure 2 is a sample of the dense encoder unit which $n = 2$. The dense decoder block is similar to the dense encoder block.

Some dense connections with special architectures in the encoder and the decoder are also interesting, such

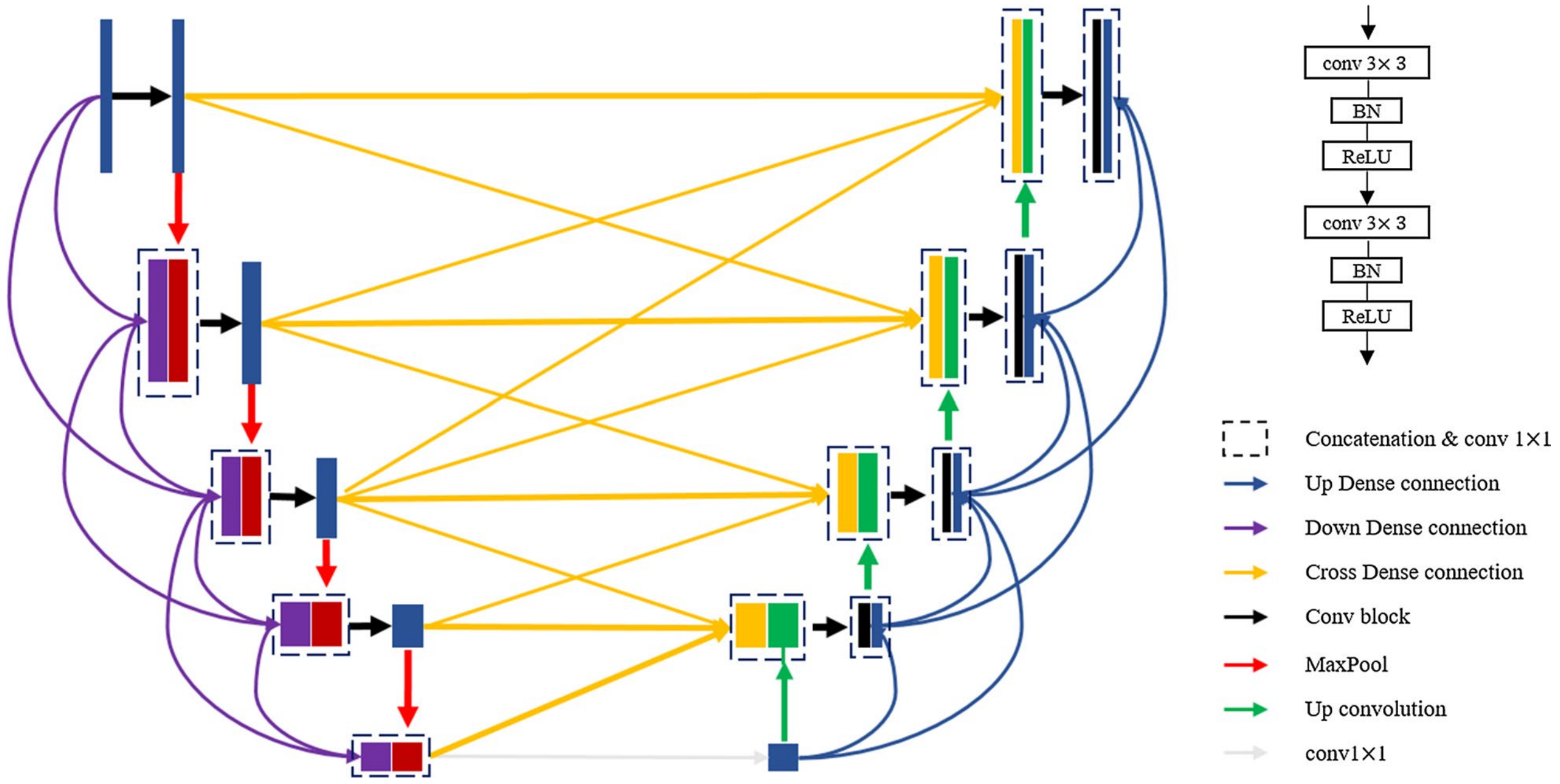

**Fig. 4** An example of the multi-scale dense connected U-Net ($encoder_2 - cross_3 - decoder_2$)

**Table 1 The comparison of U-Net with different multi-scale dense encoders**

| Method | Mean IoU | | Dice coefficient | |
|---|---|---|---|---|
| | A | B | A | B |
| U-Net | 0.797 | 0.738 | 0.886 | 0.853 |
| Min | 0.841 | 0.753 | 0.906 | 0.862 |
| $Encoder_1$ | 0.852 | 0.771 | 0.915 | 0.871 |
| $Encoder_2$ | 0.856 | 0.772 | 0.918 | 0.869 |
| $Encoder_3$ | 0.859 | **0.779** | 0.919 | **0.877** |
| $Encoder_4$ | **0.861** | 0.778 | **0.919** | 0.872 |

Best results are given in bold

**Table 2 The comparison of U-Net with different multi-scale dense decoders**

| Method | Mean IoU | | Dice coefficient | |
|---|---|---|---|---|
| | A | B | A | B |
| U-Net | 0.797 | 0.738 | 0.886 | 0.853 |
| Mout | 0.841 | 0.759 | 0.908 | 0.861 |
| $Decoder_1$ | 0.852 | 0.768 | 0.915 | 0.866 |
| $Decoder_2$ | 0.857 | 0.770 | 0.917 | 0.870 |
| $Decoder_3$ | 0.860 | 0.784 | 0.919 | **0.877** |
| $Decoder_4$ | **0.861** | **0.784** | **0.920** | 0.870 |

Best results are given in bold

as the multi-input connections (Min) and multi-output connections(Mout), which are shown in Eqs. 5 and 6. In Min, each layer only fuses the original input, which is down-sampled to the corresponding size. Meanwhile, in Mout, only the last layer fuses all the feature maps from the previous layer, which are also up-sampled to the corresponding size.

$$X_e^{i+1} = H_{min}(X_{new}^1) \tag{5}$$

$$Y_e^5 = H_{mout}(Y_{new}^1, Y_{new}^2, Y_{new}^3, Y_{new}^4) \tag{6}$$

Secondly, we introduce the multi-scale dense connections across the encoder and the decoder. We also start from the architecture of the traditional U-Net. As shown in Fig. 3, the traditional cross connections are defined as Eqs. 7, 8, and 9. $Y^{i-1}$ and $Y^i$ are the input and output of the current layer, respectively. $X^{i-1}$ is the feature map in encoder with corresponding size of $Y^i$. $Y_p^{i-1}$ is the output of $Y^{i-1}$ after up-sampling. $Y_c^{i-1}$ encoder the feature maps from layer $I-1$ in the encoder and the output from the previous layer in the decoder after up-sampling.

**Table 3 The comparison of U-Net with different multi-scale dense cross connections**

| Method | Mean IoU | | Dice coefficient | |
|---|---|---|---|---|
| | A | B | A | B |
| U-Net | 0.797 | 0.738 | 0.886 | 0.853 |
| Upper | 0.852 | 0.762 | 0.917 | 0.866 |
| Lower | 0.855 | 0.766 | 0.918 | 0.870 |
| $Cross_3$ | 0.857 | 0.770 | 0.916 | 0.868 |
| $Cross_5$ | **0.861** | **0.778** | **0.920** | **0.872** |

Best results are given in bold

**Table 4 The comparison of U-Net with different multi-scale dense connections**

| Method | Mean IoU | | Dice coefficient | |
|---|---|---|---|---|
| | A | B | A | B |
| U-Net | 0.797 | 0.738 | 0.902 | 0.842 |
| U-Net-$encoder_4 - cross_5 - \varnothing$ | 0.853 | 0.764 | 0.916 | 0.864 |
| U-Net-$encoder_4 - \varnothing - decoder_4$ | 0.859 | 0.770 | 0.918 | 0.870 |
| U-Net-$\varnothing - cross_5 - decoder_4$ | 0.863 | **0.768** | 0.920 | **0.871** |
| U-Net-$encoder_4 - cross_5 - decoder_4$ | **0.866** | 0.764 | **0.925** | 0.857 |

Best results are given in bold

$$Y_p^{i-1} = U(Y^{i-1}) \tag{7}$$

$$Y_c^i = H(X^{i-1}, Y_p^{i-1}) \tag{8}$$

$$Y^i = F(Y_c^{i-1}) \tag{9}$$

our method uses $Y_{newc}^{i-1}$ instead of $Y_c^{i-1}$ , which is defined as Eq. 11. $\widetilde{X}^{i-1}$ encode the coarse-to-fine context information from the encoder. $Y_{newc}$ fuses two feature maps $\widetilde{X}^{i-1}$ and $Y_p^{i-1}$. $H()$ represents the same operation. which adjusts the number of channels the same as $X^{i-1}$.

$$\widetilde{X}^{i-1} = H(X_{new}^{(i-n)}, \cdots, X^i, \cdots, X_{new}^{(i+n)}) \tag{10}$$

$$Y_{newc}^i = H(\widetilde{X}^{i-1}, Y_p^{i-1}) \tag{11}$$

Some dense connections across the encoder and the decoder with special architectures are also interesting, such as the Upper and the Lower, which are shown in Eqs. 12 and 13. In Upper, each layer in the decoder can only fuse the feature with higher resolutions in the encoder, while in Lower, each layer in the decoder can only fuse the feature with lower resolutions in the encoder.

$$\widetilde{X}^{i-1}_{Upper} = H(X^{(i-d)}_{new}, \cdots, X^{i}_{new}) \quad (12)$$

$$\widetilde{X}^{i-1}_{Lower} = H\left(X^{i}_{new}, \cdots, X^{(i+d)}_{new}\right) \quad (13)$$

### Multi-scale densely connected U-Net

In this section, we introduce the fully dense connected U-shape architecture based on U-Net. As illustrated in Fig. 4, the improved structure of the encoder is identical to "Dense connections" section. The decode structure is the combination of multi-scale dense cross connections and multi-scale dense decoder. The variants and operations share the same description with "Multi-scale densely connected U-Net" section.

The detailed information follows Eqs. 7, 8, and 9 in "Multi-scale densely connected U-Net" section. The variants and operations share the same description as "Multi-scale densely connected U-Net" section.

$$\widetilde{X}^{i-1} = H\left(X^{(i-n)}_{new}, \cdots, X^{i}, \cdots, X^{(i+n)}_{new}\right) \quad (14)$$

$$Y^{i-1}_{e} = H\left(Y^{((i+1)}_{new}, \cdots, Y^{(i+n)}_{new}\right) \quad (15)$$

$$Y^{i}_{ee} = H\left(\widetilde{X}^{i-1}, Y^{i-1}_{e}\right) \quad (16)$$

$$Y^{i}_{newc} = H\left(\widetilde{Y}^{i-1}_{p}, Y^{i-1}_{ee}\right) \quad (17)$$

MDU-Net encodes the dense cross connections and the dense decoder to further improve the multi-scale feature representative. We re-encode the information obtained from the first encoding operation. The encoded feature maps share the same number of channels as the original one.

**Table 5 The efficiency comparison of MDU-Net with other multi-scale densely connected networks**

| Method | Parameter |
|---|---|
| U-Net | 8M |
| U + dense encoder block | 8M + 0.005M |
| U + dense decoder block | 8M + 0.005M |
| U + dense cross connections | 8M + 0.005M |
| MDU-Net[a] | 8M + 0.015M |
| U-Net++ | 8M + 1M |
| MILDnet | 8M + 68M |
| MIMOnet | 8M + 166M |

[a] MDU-Net means that the framework contains three dense connections based on U-Net

**Table 6 The comparison of MDU-Net with different quantization strategies**

| Method | Mean IoU | | Dice coefficient | |
|---|---|---|---|---|
| | Part A | Part B | Part A | Part B |
| MDU-Net | 0.866 | 0.764 | 0.925 | 0.857 |
| MDU-Net +$INQ3_{1/2}$[a] | 0.871 | 0.784 | 0.925 | 0.873 |
| MDU-Net+$INQ3_{3/4}$ | 0.866 | 0.790 | 0.923 | 0.876 |
| MDU-Net+$INQ3_{1}$ | 0.859 | 0.791 | 0.918 | 0.865 |
| MDU-Net+$INQ5_{1/2}$ | **0.872** | 0.772 | **0.928** | 0.878 |
| MDU-Net+$INQ5_{3/4}$ | 0.865 | **0.786** | 0.922 | 0.876 |
| MDU-Net+$INQ5_{1}$ | 0.857 | 0.750 | 0.916 | **0.881** |
| MDU-Net+$INQ7_{1/2}$ | 0.867 | 0.776 | 0.919 | 0.871 |
| MDU-Net+$INQ7_{3/4}$ | 0.862 | 0.772 | 0.925 | 0.870 |
| MDU-Net+$INQ7_{1}$ | 0.859 | 0.768 | 0.922 | 0.878 |

Best results are given in bold

[a] The subscript 1/2 means that 1/2 parameters of the model are quantized

## Experiments

### Experiment setup

In this paper, we applied the Gland Segmentation (GlaS) dataset to evaluate the proposed model. GlaS is a histology image dataset published in MICCAI'2015. It contains 165 images with 16 H & E stained histological sections of colon cancer. 85 images (37 benign and 48 malignant) are selected as training sets while 80 images (37 benign and 43 malignant) are used for testing. Particularly, all test images were separated into two categories (60 Test Part A and 20 Test Part B). We train our proposed end-to-end network with back-propagation on two NVIDIA GeForce GTX TITAN X, each containing 12 GB of memory. We set the learning rate to 0.005 in the beginning and divide by 10 when the iteration reaches a threshold. SGD optimization algorithm and a batch size of 4 are set during the training. Additionally, we conduct experiments on dense connections of various sizes and shapes. For dense encoder and dense decoder, we compare the base model with a different number of connections (from 1 to 4) and two special cases (Min and Mout). For dense cross, we validate the effectiveness of two special cases ($cross_3$ and $cross_5$). Besides, we also analyze the impact of the network of quantization. We adopt Incremental Quantization (INQ) [44] as a base quantization method to compress the weights. We conduct experiments on different bits of 3, 5, and 7 to reduce the potential overfitting of dense connections, and the experimental results are shown in "Impact of network quantization" section.

### Quantitative results

In this section, we explore the impact of three dense structures (dense encoder, dense decoder, dense cross) with the number of connections varying in detail.

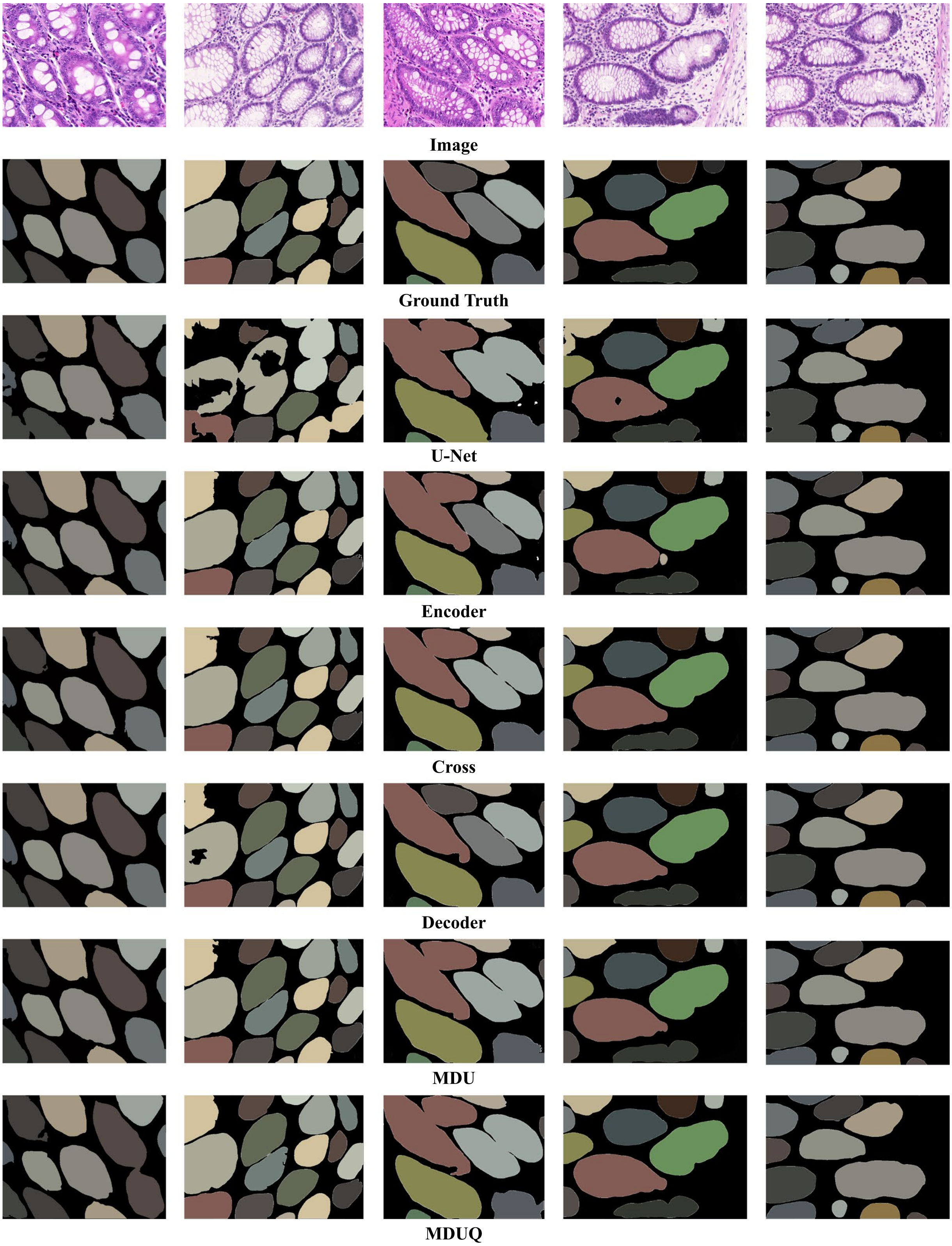

**Fig. 5** The visualization results on the GlaS dataset

Tables 1, 2, and 3 demonstrate the impact of three dense structures with different connection numbers. The experimental results show that the accuracy is generally improved as the number of dense connections increases. On MICCAI 2015 GlaS dataset, the modified structures achieves a superiority by 2% on average over U-Net. The experimental result indicates dense connections including the encoded object information from higher layers and pixel information from lower layers improve the feature reuse and thus gain a promising segmentation accuracy. In terms of different locations of dense connections, we find that the improvement of dense connections in the decoder is slightly larger than the other two, which may be due to the fact that multi-scale features in the decoder contain the global and local features, and thus their fusion can improve the overall performance more than the other two.

Furthermore, we investigate the impact of combining three different dense connected blocks. We have reached a conclusion before that the increasing number of dense connections results in a better performance of the model. We select $encoder_4$ as the basic component, indicating feature maps in each encoding block contribute to four subsequent blocks and $decoder_4$ is chosen in the same manner. Note that we set $cross_5$ connections consisting of two upper connections from subsequent layers, two lower connections from previous layers, and the direct skip connection as U-Net. We systematically conduct the experiment of combining two or three basic components. The result is shown in Table 4. Obviously, in Test A, either the combination of two or three achieves a reasonable improvement. However, in Test B, the performance drops compared with the single model. We believe the decreased accuracy is caused by the potential overfitting as the distribution of the train dataset and test set A is approximately closer. Figure 5 shows the visual results from the models with different multi-scale dense connections.

### Analysis of network efficiency

Apart from assessing the accuracy of segmentation, we evaluate the efficiency of the proposed network. Recent methods based on U-Net appear wider, deeper, and thus more difficult to optimize. In contrast, even the fully dense connected structure only increases a tiny number of parameters compared with U-Net. Because there are no extra computations and parameters involved except for the $1 \times 1$ convolutional layers in the feature reuse and concatenation operation. Table 1 demonstrates the comparison of the number of parameters of existing methods. We achieve state-of-the-art accuracy while exhibiting an ignorable increment of parameters, which reveals the high efficiency of our proposed model. On the other hand, our proposed model reveals a valuable extendibility and can be treated as a novel backbone rather than U-Net for U-shape-based networks (see Table 5).

### Impact of network quantization

In this section, we explore using quantization methods to improve the performance of our proposed network. In particular, Incremented Quantization [44] is applied as the basic method to quantize the weights. Note that quantizing the full-precision weights may reduce potential overfitting, but fully completely quantizing all the weights often leads to a reduction in segmentation accuracy. Thus, we validate the intermediate results of the quantization step in anticipation of a trade-off between the mitigation of overfitting and the loss of accuracy brought about by model quantization. As stated in Table 6, the overfitting problem is largely reduced after the first quantization operation in which half of the weights are quantized. However, after the weights are fully quantized, the model has a decrease in segmentation performance compared to the model whose weights are partially quantized. This may be due to the fact that full quantization is an aggressive quantization method, which may hurts the segmentation performance of the model. We obtain a surprising accuracy of 0.88 on test B. In balance, we adopt the half-quantized architecture as our final model.

## Conclusion

In this paper, we propose three different multi-scale dense connections for U-shaped architecture's encoder, decoder, and across them. Our architecture directly fuses the neighboring feature maps with different scales from both higher layers and lower layers to strengthen feature propagation in the current layer, which can largely enhance the feature aggregation in the encoder, the decoder, and across them. And next, we explore their combinations in detail based on U-Net. The experimental results shows that the accuracy generally improved with the number of dense connections increases. We adopt the optimal model based on the experiment and propose a novel MDU-Net combining three dense connected architectures with quantization, which reduces the overfitting from dense connections. Finally, our MDU-Net achieves the superiority dice coefficient over U-Net by up to 3% on test A and 4.1% on test B.

**Acknowledgements**
This work was supported by the Major Key Project of PCL (Grant Nos. PCL2022A03, PCL2021A02, PCL2021A09), the Natural Science Foundation of Guangdong Province (Grant No. 2022A1515010157), the National Key Research and Development Program of China (Grant No. 2018YFC1002600), the Science and Technology Planning Project of Guangdong Province, China (Grant Nos. 2017B090904034, 2017B030314109, 2018B090944002,

2019B020230003), Guangdong Peak Project (Grant No. DFJH201802), and the National Natural Science Foundation of China (Grant No. 62006050).

**Author details**
[1]The Department of New Networks, Peng Cheng Laboratory, Shenzhen, Guangdong, China. [2]Department of Cardiovascular Surgery, Guangdong Provincial People's Hospital (Guangdong Academy of Medical Sciences),Southern Medical University, Guangzhou, Guangdong, China. [3]Cyberspace Institute of Advanced Technology, Guangzhou University, Guangzhou, Guangdong, China. [4]Institute for Sustainable Industries & Livable Cities, Victoria University, Melbourne, VIC, Australia. [5]Shanghai Key Laboratory of Data Science, School of Computer Science, Fudan University, Shanghai, China.